\documentclass{article}

\usepackage{arxiv}

\usepackage[utf8]{inputenc} 
\usepackage[T1]{fontenc}    
\usepackage{hyperref}       
\usepackage{url}            
\usepackage{booktabs}       
\usepackage{amsfonts}       
\usepackage[linesnumbered,ruled,longend]{algorithm2e}
\usepackage{microtype}      
\usepackage{subfigure}
\usepackage{placeins}
\usepackage{enumitem}
\usepackage{graphicx}
\graphicspath{{../plots/}}
\newcommand{\code}[1]{\texttt{#1}}

\title{A Heuristic For Efficient Reduction In Hidden Layer Combinations For Feedforward Neural Networks}

\author{
  W. H.~Khoong\thanks{Proceedings of the 2020 Computing Conference.} \\
  Department of Statistics and Applied Probability\\
 National University of Singapore\\
  \texttt{khoongweihao@u.nus.edu} \\
}

\date{September 27 2019}

\begin{document}
\maketitle

\begin{abstract}
In this paper, we describe the hyperparameter search problem in the field of machine learning and present a heuristic approach in an attempt to tackle it. In most learning algorithms, a set of hyperparameters must be determined before training commences. The choice of hyperparameters can affect the final model's performance significantly, but yet determining a good choice of hyperparameters is in most cases complex and consumes large amount of computing resources. In this paper, we show the differences between an exhaustive search of hyperparameters and a heuristic search, and show that there is a significant reduction in time taken to obtain the resulting model with marginal differences in evaluation metrics when compared to the benchmark case.
\end{abstract}


\section{Preliminaries}

Much research has been done in the field of hyperparameter optimization~\cite{claesen2015, feurer2015, bergstra2012}, with approaches such as grid search, random search, Bayesian optimization, gradient-based optimization, etc. Grid search and manual search are the most widely used strategies for hyperparameter optimization~\cite{bergstra2012}. These approaches leave much room for reproducibility and are impractical when there are a large number of hyperparameters. For example, gird search suffers from the curse of dimensionality when the number of hyperparameters grow very large, and manual tuning of hyperprameters require considerable expertise which often leads to poor reproducibility, especially with a large number ofhyperparameters~\cite{claesen2014}. Thus, the idea of automating hyperparameter search is increasingly being researched upon, and these automated approaches have already been shown to outperform manual search by numerous researchers across several problems~\cite{bergstra2012}.

A multilayer perceptron (MLP) is a class of feedforward artificial neural network (ANN) which can be viewed as a logistic regression classifier, where the inputs are transformed using a learnt non-linear transformation and stored in an input layer. Every element which holds an input is called a "neuron". A MLP typically consists of at least three layers of nodes: an input layer, a hidden layer and an output layer. With the exception of the input layer, each node in the layer is a neuron that utilizes a nonlinear activation function. In training the MLP, backpropagation, a supervised learning technique is used. In our experiments, we only have test cases consisting of one to three hidden layers, each consisting of up to $10$ neurons. The reasons for this number are that our objective is to illustrate the effects of the heuristic using a small toy-example that does not take too long to run in the test cases. Moreover, we found that for the datasets used, the best results from our experiments with grid search and a limit of 10 neurons in each layer involed way less than 10 neurons for each layer.

\section{Related Work}

There are related work involving search heuristics with similar methods of search, such as the comparison of the error metrics at each iteration of the algorithm. To the best of our knowledge, there has yet to be similar work which invloves the result of a grid search as input to the search algorithm, especially in the area of neural architecture search. 

Similar work was done for time series forecasting~\cite{silva2008}, where the authors employed a local search algorithm to estimate the parameters for the non-linear autoregressive integrated moving average (NARMA) model. In the paper, the algorithm's improvement method progressively searches for a new value in a small neighborhood of the underlying solution until all the parameter vector's elements have been analyzed. The error metric of interest was the mean square error (MSE), where a new direction of search was created when the new vector produces a smaller MSE. Another work~\cite{jordanov2007} which utilized a similar improvement method presented a hybrid global search algorithm for feedforward neural networks supervised learning, which combnies a global search heuristic on a sequence of points and a simplex local search. 

\section{Experiment Setting and Datasets}

\subsection{Programs Employed}

We made use of Scikit-Learn~\cite{scikit-learn}, a free software machine learning library for the Python programming language. Python 3.6.4 was used in formulating and running of the algorithms, plotting of results and for data preprocessing. 

\subsection{Resources Utilized}

All experiments were conducted in the university's High Performance Computing\footnote{See \url{https://nusit.nus.edu.sg/services/hpc/about-hpc/} for more details about the HPC.} (HPC) machines, where we dedicated 12 CPU cores, 5GB of RAM in the job script. All jobs were submitted via SSH through the \code{atlas8} host, which has the specifications: $\textit{HP Xeon two sockets 12-Core 64-bit Linux cluster, CentOS 6}$. We utilized \textit{Snakemake}~\cite{snakemake}, a workflow management tool to conduct our experiments. 

\subsection{Data}

We perform the experiments on two datasets - Boston house-prices and the MNIST handwritten digit dataset~\cite{mnist}. The Boston house-prices dataset\footnote{See \url{https://scikit-learn.org/stable/modules/generated/sklearn.datasets.load\_boston.html} for the documentation.} is available from Scikit-Learn's \code{sklearn.datasets} package, and . This package contains a few small standard datasets that do not require downloads of any file(s) from an external website. The MNIST dataset can be downloaded from \url{http://yann.lecun.com/exdb/mnist/} .

\subsection{Notations and Test Cases}

We perform our experiments on feedforward neural networks with one, two and three layers, with Scikit-Learn's \code{MLPRegressor} and \code{MLPClassifier} from the \code{sklearn.neural\_network}\footnote{See \url{https://scikit-learn.org/stable/modules/classes.html\#module-sklearn.neural\_network} for further documentation.} package. The \code{MLPRegressor} is used for predicting housing prices for the Boston dataset and the \code{MLPClassifier} is used for classification with the MNIST dataset. The models optimize the squared-loss using the Limited-memory Broyden–Fletcher–Goldfarb–Shanno algorithm~\cite{fletcher1987} (LBFGS), an optimizer in the family of quasi-Newton methods. The hyperparameter values were all fixed except for the number of neurons at each hidden layer, denoted by \code{hidden\_layer\_sizes}, which is dependent on the current input at each iteration of the algorithm. The other (fixed) hyperparameters are: \code{activation}(\code{relu}), \code{solver} (\code{lbfgs}), \code{alpha} (\code{0.0001}), \code{batch\_size} (\code{auto}), \code{learning\_rate} (\code{constant}), \code{learning\_rate\_init} (\code{0.001}), \code{max\_iter} (\code{500}) and \code{random\_state} (\code{69}). All other hyperparameters not stated here are using their default values. Further information about the hyperparameters can be found in the \code{sklearn.neural\_network} documentation. Our maximum number of neurons in any hidden layer is set at $10$, as preliminary experiments show that the best stratified 5-fold cross-validation score is obtained when the number of neurons at any hidden layer is under $10$. 

Define $\alpha$ to be the $\textit{minimum fraction in improvement}$ required for each iteration of the algorithm. We also define $H^{(i)}, i = 0,1,\dots$ to be the set of hidden-layer combination(s) at iteration $i$. $H^{(0)}$ is the starting set of hidden-layer combinations used as input, with $|H^{(0)}| = 1$. Let $N$ be the number of hidden layers and $H_{\textit{model}}^{(j)}$ to be the set containing the best hidden layer combination obtained from fitting with \code{GridSearchCV}\footnote{See documentation at \url{https://scikit-learn.org/stable/modules/generated/sklearn.model_selection.GridSearchCV.html}.} from Scikit-Learn's \code{sklearn.model\_selection} package. For example, if $H^{(0)} = \{(3,4,3)\}$, it means that there are $3$ neurons in the first and third hidden layer and $4$ neurons in the second layer. We also define $H_{\textit{prev}}$ as the set contianing all previously fitted hidden layer combinations. $H_{\textit{best}}$ is then the set containing the best combination at any iteration of the algorithm, i.e. $|H_{\textit{best}} | \leq 1$. Scikit-Learn's \code{sklearn.preprocessing.StandardScaler}\footnote{Further documentation can be found at \url{https://scikit-learn.org/stable/modules/generated/sklearn.preprocessing.StandardScaler.html}.} was also used to standardize features by removing the mean and scaling to unit variance. 

We also denote the Root Mean Square Error (RMSE) from fitting the model with validation on the test dataset, at the end of the current iteration and from the previous iteration as $\textit{RMSE}_{\textit{model}}, \textit{RMSE}_{\textit{curr}}$ and $\textit{RMSE}_{\textit{prev}}$ respectively. In our experiments, $\alpha \in \{0.01, 0.05, 0.10\}$. We also set the initial upper-bound threshold $K$ on the RMSE to be an arbitrarily large, for the purpose of passing the first iteration of the loop. Next, we define $\textit{Combinations}(\cdot)$ as a function that returns the set of all possible hidden layers $L_{hls}$ without duplicates.

\section{Methods Employed}
\label{sec:methods}

\subsection{Method 1 - Benchmark}

In this method, all possible hidden-layer sizes (with repetition allowed) are used as hyperparameter. Let $L_{hls}$ denote the set of all possible hidden layers. Then for example, if there are 2 hidden layers and each layer can have between 1 to 3 neurons, then $L_{hls} = \{(1,1),(1,2),(1,3),(2,1),(2,2),(2,3),(3,3),(3,2),(3,1)\}$. 

\subsection{Method 2 - Heuristic}

In this method, a heuristic is used to iteratively explore the hidden-layer combinations, subject to the condition that the abosolute change in $\textit{RMSE}$ is greater or equal to $\alpha$ and that $\textit{RMSE}_{\textit{curr}} > \textit{RMSE}_{\textit{prev}}$. To be precise, the algorithm optimizes within a fixed numberof hidden layers. In our experiments, we obtain the input $H^{(0)}$ by performing a grid search on an initialization of hidden-layer combinations of the form: $L_0 = \{(1, 1, \dots, 1), \dots, (10, 10, \dots, 10)\}$ and the `best' hidden-layer combination $\{(i, i, \dots, i)\}, i \in \{1, \dots, 10\}$ will be assigned as $H^{(0)}$. Let $n_{\textit{max}}^{(\textit{input})}$ to be the maximum number of neurons across all hidden layers in $L_0$. In the above example, $n_{\textit{max}}^{(\textit{input})} = 10$. The sequence of steps of the heuristic and its pseudocode are as follows:
\begin{enumerate}
	\item Initialize a set of hidden-layer combinations of the form: $L_0 = \{(1, 1, \dots, 1), \dots, (i, i, \dots, i), \dots\}, i \geq 1$. Also set an arbitrary large value for $\textit{RMSE}_{prev}$ and $\textit{RMSE}_{curr}$.
	\item Perform a grid search with $L_0$ to obtain $H^{(0)}$, which corresponds to the set that contains the combination with the lowest test set RMSE from  grid search with stratified K-fold cross-validaton.
	\item Generate the current iteration's set of hidden-layer combinations without duplicates, with:
	\begin{enumerate}[label=(\alph*).]
		\item $n_{\textit{min}}$ as the minimum number of neurons in any hidden-layer combination in the previous iteration's set of hidden-layer combinations, deducting the current iteration's index. If $n_{\textit{min}} < 1$, set it to $1$.
		\item $n_{\textit{max}}$ as the maximum number of neurons in any hidden-layer combination in the current iteration's set of hidden-layer combinations, with an increment of $1$. If $n_{\textit{max}} > n_{\textit{max}}^{(\textit{input})}$, set $n_{\textit{max}}$ as $n_{\textit{max}}^{(\textit{input})}$. 
	\end{enumerate}
    \item If the set of hidden-layer combinations for the current iteration is empty, the algorithm terminates. Otherwise, obtain the best hidden-layer combination of the set, and set it as the current iteration's best combination. Update the iteration's set of hidden-layer combinations to the set of previously fitted hidden-layer combinations and the current iteration's best combination as the overall best hidden-layer combination.  
    \item Repeat steps 3 and 4. If the algorithm terminates in as a consequence of step 4, return the last found best hidden-layer combination.
\end{enumerate}

\begin{algorithm}[H]
	\SetAlgoLined
	\DontPrintSemicolon
	\SetKwInOut{Input}{Input}
	\SetKwInOut{Output}{Output}
	\Input{$\alpha, H^{(0)}, n_{\textit{max}}^{(\textit{input})}, N, K, \textit{Combinations}(\cdot)$}
	\Output{$\textit{RMSE}_{\textit{curr}}, H_{\textit{best}}$}
	$H_{\textit{prev}} \leftarrow \{\} $\;
	$H_{\textit{best}} \leftarrow \{\} $\;
	$\textit{RMSE}_{\textit{prev}} \leftarrow K$ \;
	$\textit{RMSE}_{\textit{curr}} \leftarrow K^2$ \;
	$i \leftarrow 0$\;
	$\Delta = {\bigg |}\frac{\textit{RMSE}_{\textit{curr}} - \textit{RMSE}_{\textit{prev}}}{\textit{RMSE}_{\textit{prev}}} {\bigg |}$\;
	\While{$\Delta \geq \alpha$ \textbf{and} $\textit{RMSE}_{\textit{curr}} > \textit{RMSE}_{\textit{prev}}$}{
		$\textit{RMSE}_{\textit{prev}} \leftarrow \textit{RMSE}_{\textit{curr}}$\;
		\If{$i = 0$}{
			$H^{(i+1)} \leftarrow H^{(0)}$\;
		}
		\Else{
			$n_{\textit{min}} \leftarrow \min(H^{(i)})-i$\;
			\If{$n_{\textit{min}} < 1$}{
				$n_{\textit{min}} \leftarrow 1$\;
			}
			$n_{\textit{max}} \leftarrow \max (H^{(i)} ) + 1$\;
			\If{$n_{\textit{max}} > n_{\textit{max}}^{\textit{(input)}}$}{
				$n_{\textit{max}} \leftarrow n_{\textit{max}}^{\textit{(input)}}$\;
			}
			$H^{(i+1)} \leftarrow \textit{Combinations} (n_{\textit{min}} , n_{\textit{max}}, N)$\;
		}
		\If{$\textit{length} (H^{(i+1)} ) = 0$}{
			\textbf{Break}\;
		}
		$\textit{model} \leftarrow \textit{fit} (H^{(i+1)})$\;
		$H_{\textit{curr}, \textit{best}} \leftarrow H_{\textit{model}}^{(i+1)}$\;
		$H_{\textit{prev}} \leftarrow H_{\textit{prev}} \cup H^{(i+1)} \setminus H^{(i+1)} \cap H_{\textit{prev}}$\;
		\If{$\textit{length} (H_{\textit{best}}) = 0$}{
			$H_{\textit{best}} \leftarrow H_{\textit{best}} \cup H_{\textit{curr}, \textit{best}}$\;
		}
		\Else{
			$H_{\textit{best}} \leftarrow \{ \}$\;
			$H_{\textit{best}} \leftarrow H_{\textit{best}} \cup H_{\textit{curr}, \textit{best}}$\;
		}
		\If{$i = 0$}{
			$\textit{RMSE}_{\textit{prev}} \leftarrow \frac{1}{2} \textit{RMSE}_{\textit{model}}$\;
		}
		$\textit{RMSE}_{\textit{curr}} \leftarrow \textit{RMSE}_{\textit{model}}$\;
		$i \leftarrow i + 1$\;
	}
	\textbf{Return} $\textit{RMSE}_{\textit{curr}}, H_{\textit{best}}$
	\caption{Heuristic}
\end{algorithm}

\section{Experiment Results}

For each dataset, we illustrate the results of Method 1 (Benchmark) and Method 2 (Heuristic) for each $\alpha$ side-by-side, then show the overall results in a table. The time elapsed for Method 2 is the time taken to perform the grid search to obtain the initial input and to obtain the result from the algorithm with the initial input from the grid search's result. For Method 1, it is the time taken to perform a grid search on all possible combinations of neurons to obtain the best result. 

\subsection{Boston Dataset}

\subsubsection{Method 1}

\begin{figure}[h]
	\hfill
	\subfigure[Score]{\includegraphics[width=0.32\textwidth]{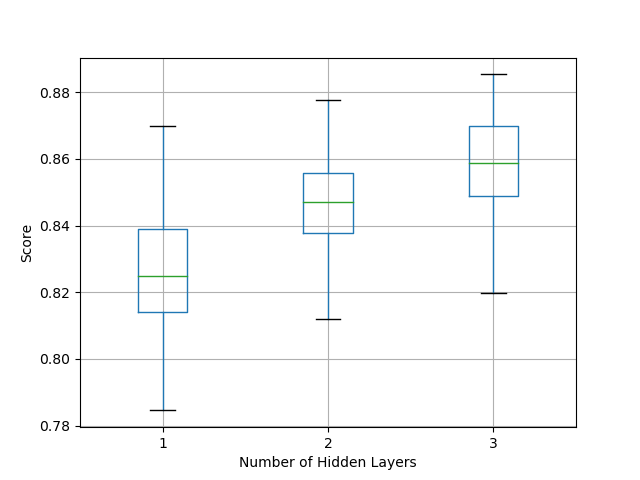}}
	\hfill
	\subfigure[Test RMSE]{\includegraphics[width=0.32\textwidth]{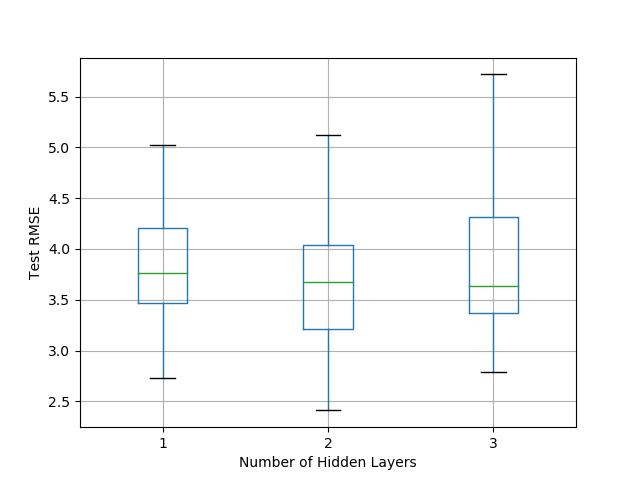}}
	\hfill
	\subfigure[Time Elapsed]{\includegraphics[width=0.32\textwidth]{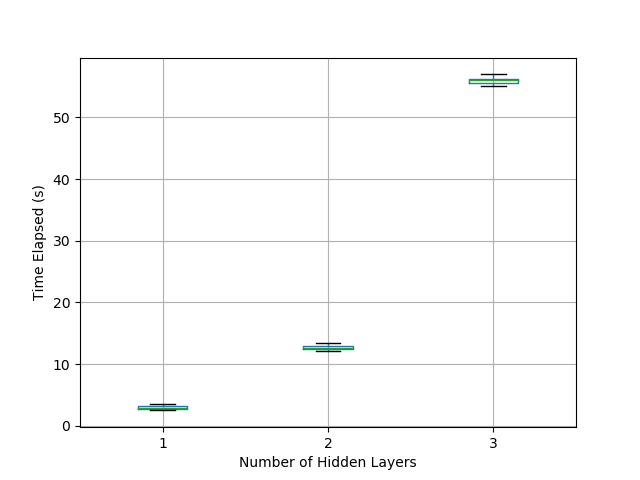}}
	\hfill
	\caption{Method 1 Results}
\end{figure}
\FloatBarrier

\subsubsection{Method 2}

\begin{figure}[h]
	\hfill
	\subfigure[Score]{\includegraphics[width=0.32\textwidth]{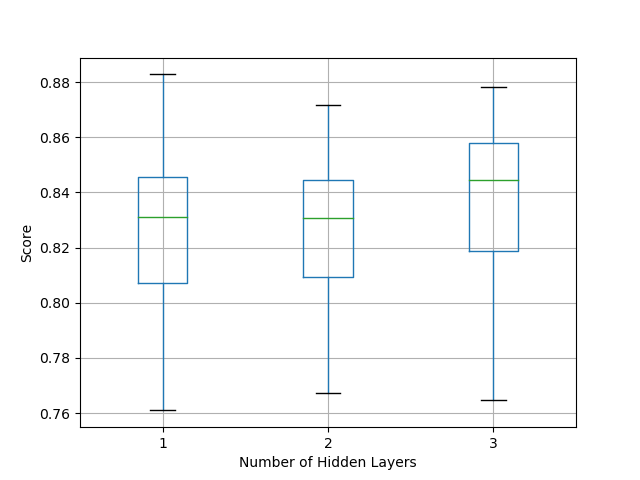}}
	\hfill
	\subfigure[Test RMSE]{\includegraphics[width=0.32\textwidth]{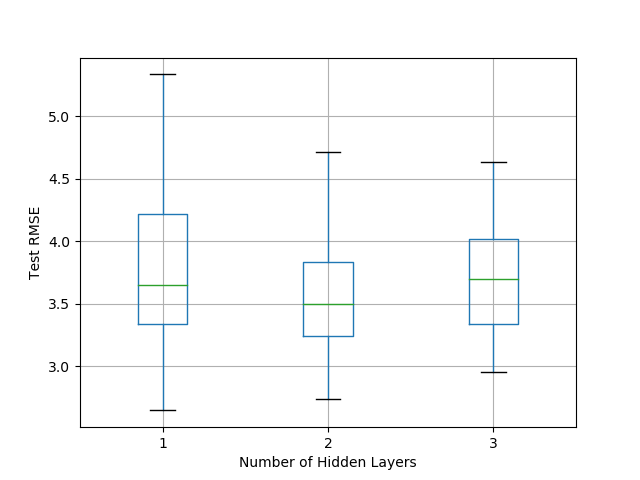}}
	\hfill
	\subfigure[Time Elapsed]{\includegraphics[width=0.32\textwidth]{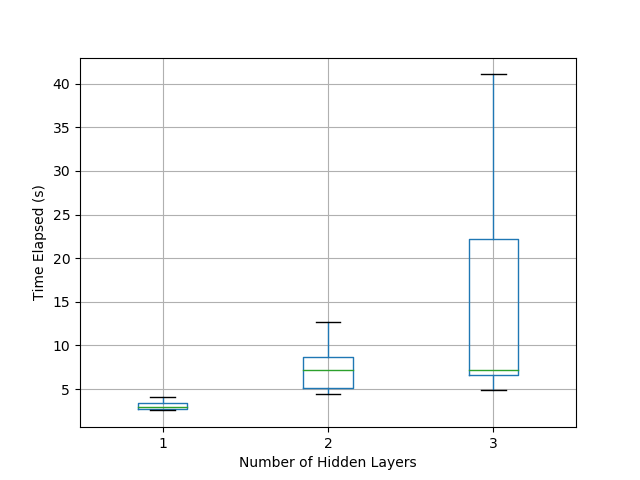}}
	\hfill
	\caption{Method 2 Results, $\alpha=0.01$}
\end{figure}
\FloatBarrier

\begin{figure}[h]
	\hfill
	\subfigure[Score]{\includegraphics[width=0.32\textwidth]{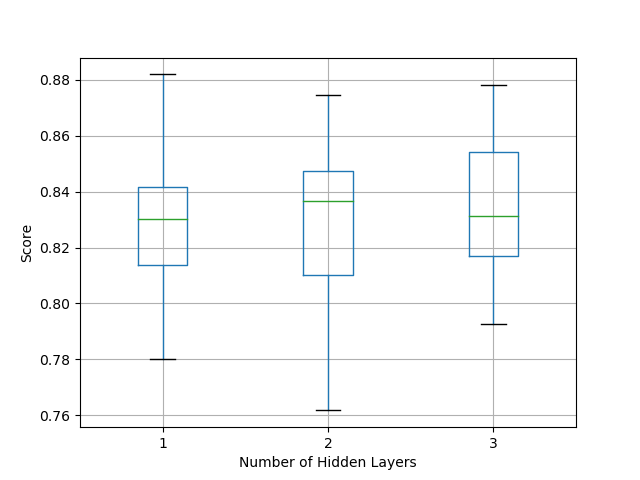}}
	\hfill
	\subfigure[Test RMSE]{\includegraphics[width=0.32\textwidth]{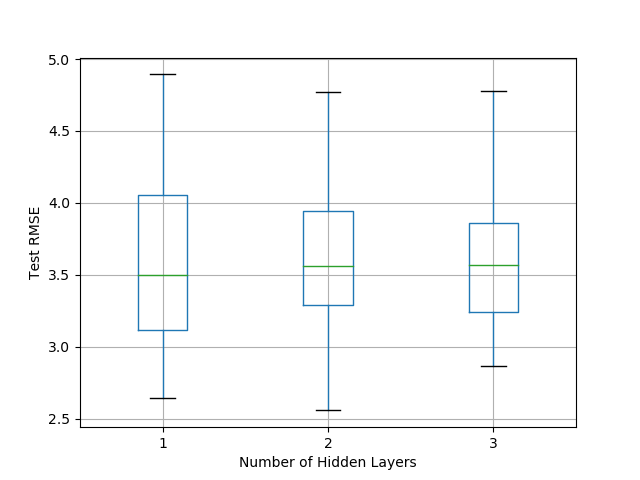}}
	\hfill
	\subfigure[Time Elapsed]{\includegraphics[width=0.32\textwidth]{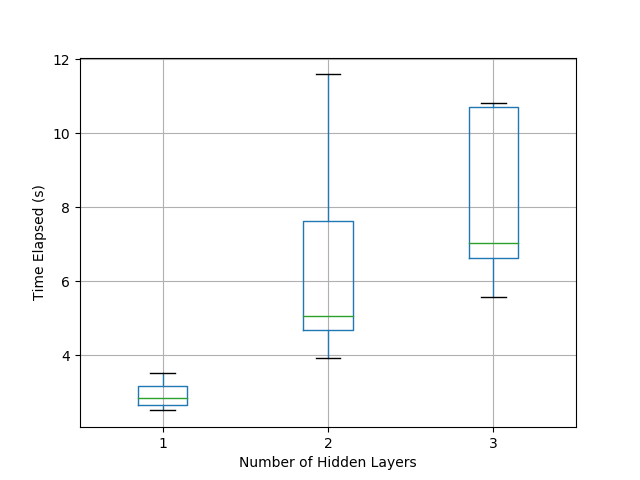}}
	\hfill
	\caption{Method 2 Results, $\alpha=0.05$}
\end{figure}
\FloatBarrier

\begin{figure}[h]
	\hfill
	\subfigure[Score]{\includegraphics[width=0.32\textwidth]{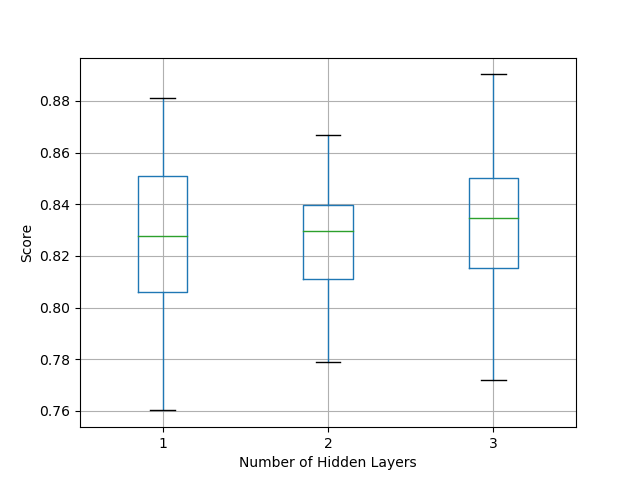}}
	\hfill
	\subfigure[Test RMSE]{\includegraphics[width=0.32\textwidth]{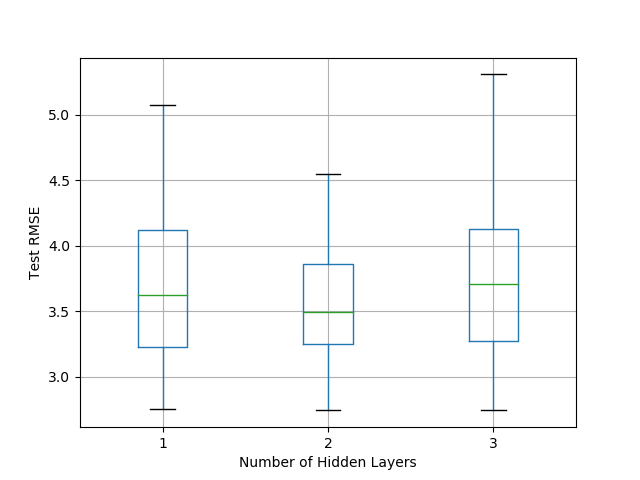}}
	\hfill
	\subfigure[Time Elapsed]{\includegraphics[width=0.32\textwidth]{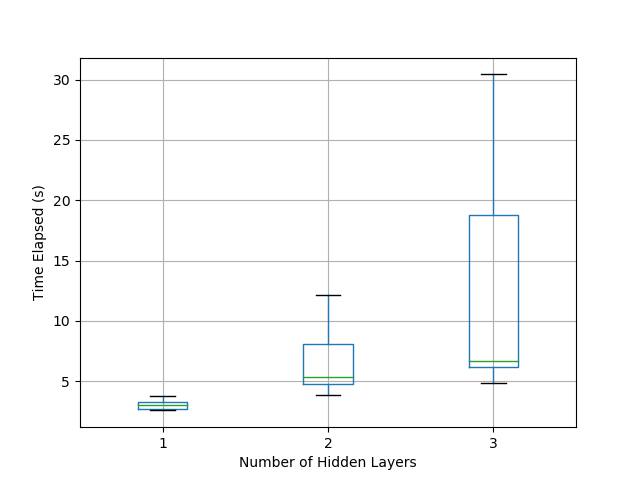}}
	\hfill
	\caption{Method 2 Results, $\alpha=0.10$}
\end{figure}
\FloatBarrier

\subsubsection{Summary of Results}

\begin{table}[!htbp]
	\centering
	\begin{tabular}{|c|c|c|c|}
		\hline
		& 1 & 2 & 3 \\ \hline
		\textbf{Median Score} & 0.83 & 0.85 & 0.86 \\ \hline
		\textbf{Median RMSE} & 3.76 & 3.68 & 3.63 \\ \hline
		\textbf{Median Time Elapsed (s)} & 2.92 & 12.63 & 56.00 \\ \hline
	\end{tabular}
	\caption{Summary of Results for Method 1}
	\label{benchmark}
\end{table}
\FloatBarrier

\begin{table}[!htbp]
	\centering
	\begin{tabular}{|c|c|c|c|}
		\hline
		& 1 & 2 & 3 \\ \hline
		\textbf{Median Score} & 0.83 & 0.83 & 0.84 \\ \hline
		\textbf{Median RMSE} & 3.65 & 3.50 & 3.70 \\ \hline
		\textbf{Median Time Elapsed (s)} & 2.95 & 7.19 & 7.22 \\ \hline
	\end{tabular}
	\caption{Summary of Results for Method 2, $\alpha= 0.01$}
	\label{alpha0.01}
\end{table}
\FloatBarrier

\begin{table}[!htbp]
	\centering
	\begin{tabular}{|c|c|c|c|}
		\hline
		& 1 & 2 & 3 \\ \hline
		\textbf{Median Score} & 0.83 & 0.84 & 0.83 \\ \hline
		\textbf{Median RMSE} & 3.50 & 3.56 & 3.57 \\ \hline
		\textbf{Median Time Elapsed (s)} & 2.84 & 5.07 & 7.03 \\ \hline
	\end{tabular}
	\caption{Summary of Results for Method 2, $\alpha= 0.05$}
	\label{alpha0.05}
\end{table}
\FloatBarrier

\begin{table}[!htbp]
	\centering
	\begin{tabular}{|c|c|c|c|}
		\hline
		& 1 & 2 & 3 \\ \hline
		\textbf{Median Score} & 0.83 & 0.83 & 0.83 \\ \hline
		\textbf{Median RMSE} & 3.62 & 3.49 & 3.71 \\ \hline
		\textbf{Median Time Elapsed (s)} & 3.04 & 5.32 & 6.69 \\ \hline
	\end{tabular}
	\caption{Summary of Results for Method 2, $\alpha= 0.10$}
	\label{alpha0.10}
\end{table}
\FloatBarrier

\newpage
\subsection{MNIST Dataset}

\subsubsection{Method 1}

\begin{figure}[h]
	\hfill
	\subfigure[Score]{\includegraphics[width=0.32\textwidth]{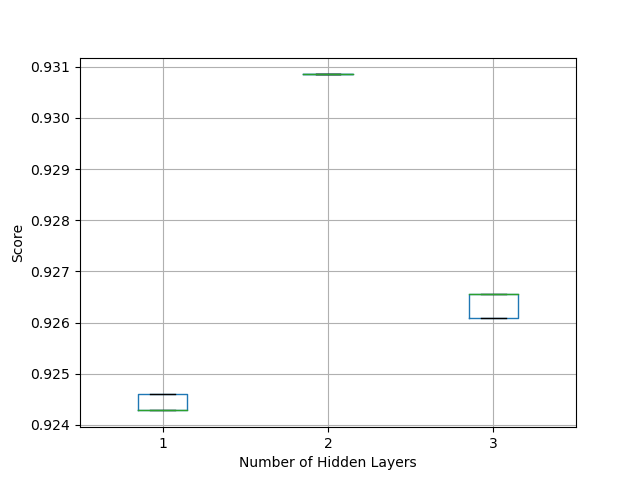}}
	\hfill
	\subfigure[Test RMSE]{\includegraphics[width=0.32\textwidth]{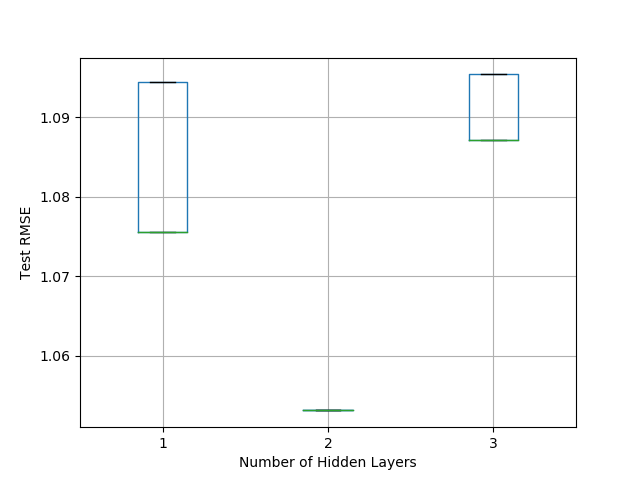}}
	\hfill
	\subfigure[Time Elapsed]{\includegraphics[width=0.32\textwidth]{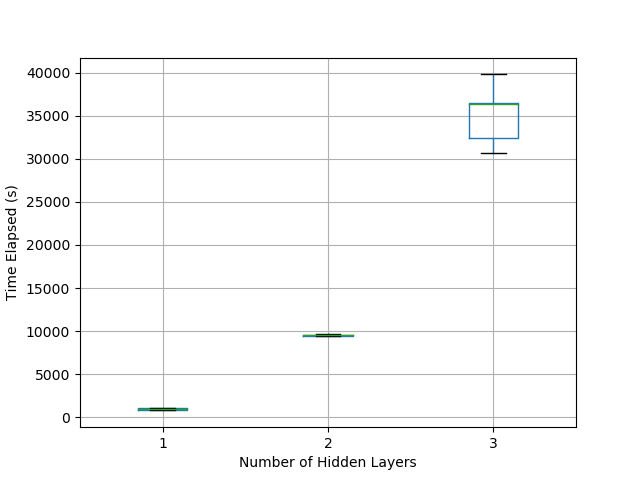}}
	\hfill
	\caption{Method 1 Results}
\end{figure}
\FloatBarrier

\subsubsection{Method 2}

\begin{figure}[h]
	\hfill
	\subfigure[Score]{\includegraphics[width=0.32\textwidth]{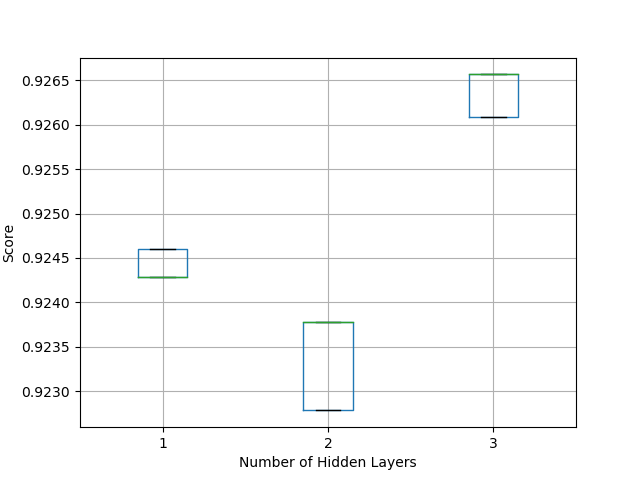}}
	\hfill
	\subfigure[Test RMSE]{\includegraphics[width=0.32\textwidth]{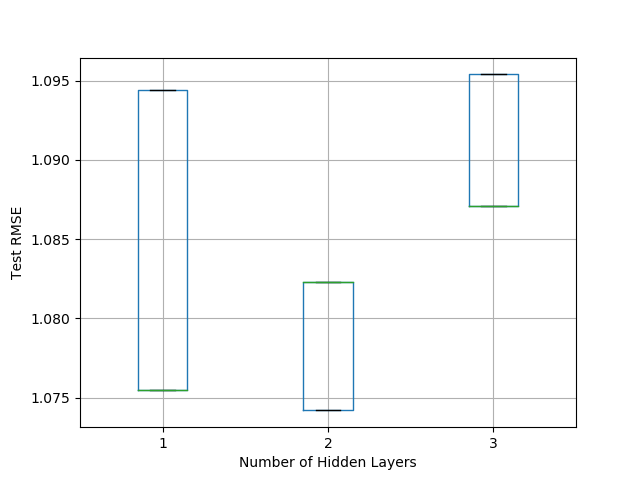}}
	\hfill
	\subfigure[Time Elapsed]{\includegraphics[width=0.32\textwidth]{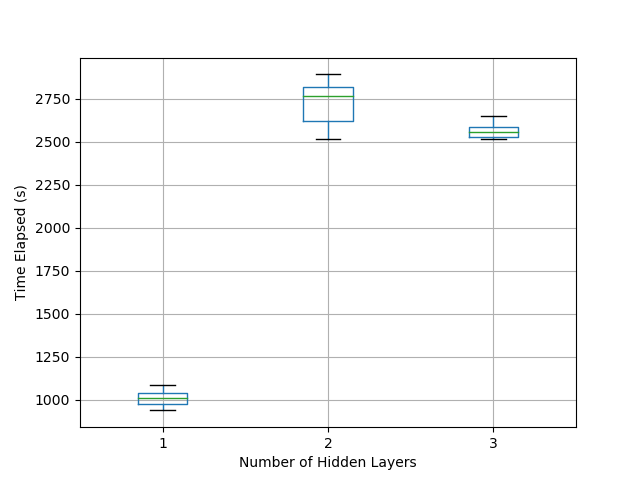}}
	\hfill
	\caption{Method 2 Results, $\alpha=0.01$}
\end{figure}
\FloatBarrier

\begin{figure}[h]
	\hfill
	\subfigure[Score]{\includegraphics[width=0.32\textwidth]{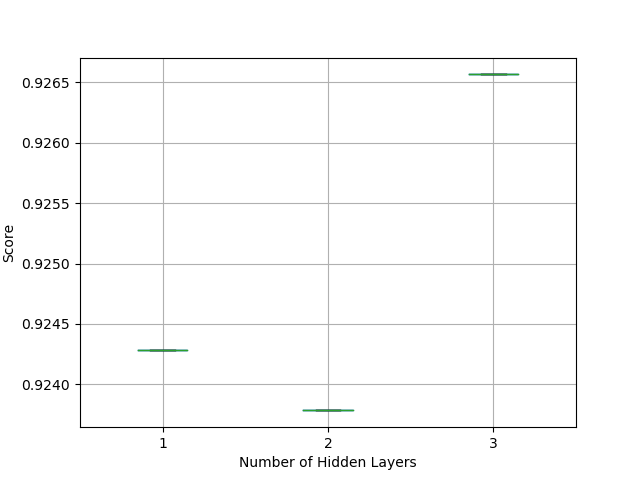}}
	\hfill
	\subfigure[Test RMSE]{\includegraphics[width=0.32\textwidth]{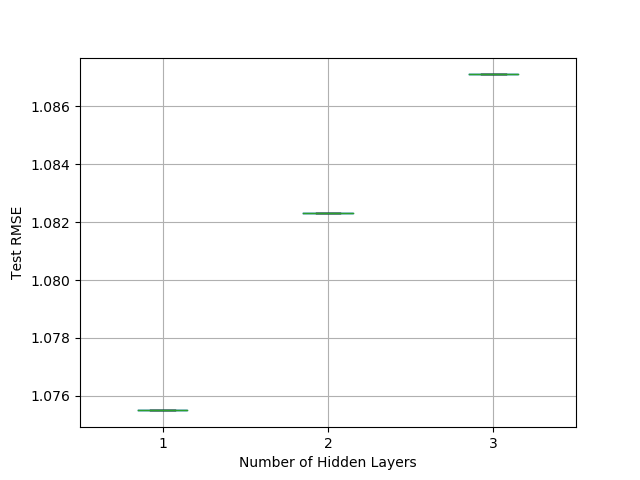}}
	\hfill
	\subfigure[Time Elapsed]{\includegraphics[width=0.32\textwidth]{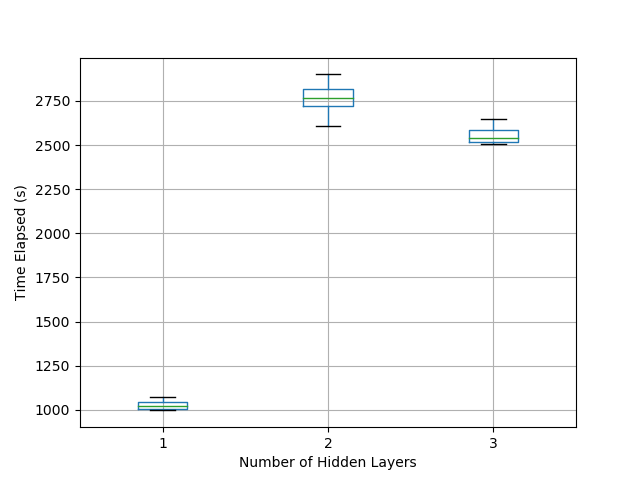}}
	\hfill
	\caption{Method 2 Results, $\alpha=0.05$}
\end{figure}
\FloatBarrier

\begin{figure}[h]
	\hfill
	\subfigure[Score]{\includegraphics[width=0.32\textwidth]{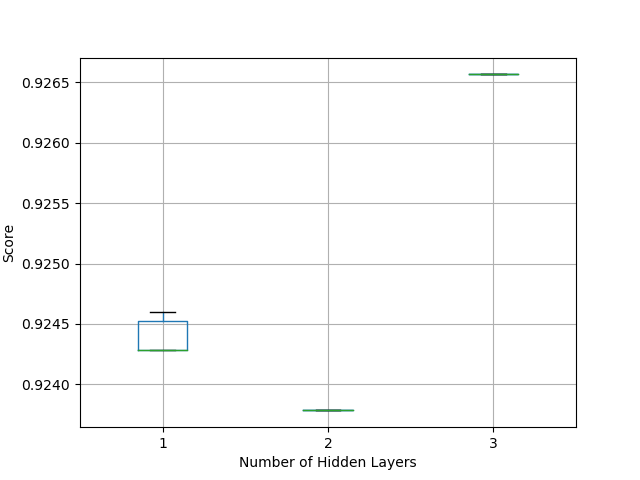}}
	\hfill
	\subfigure[Test RMSE]{\includegraphics[width=0.32\textwidth]{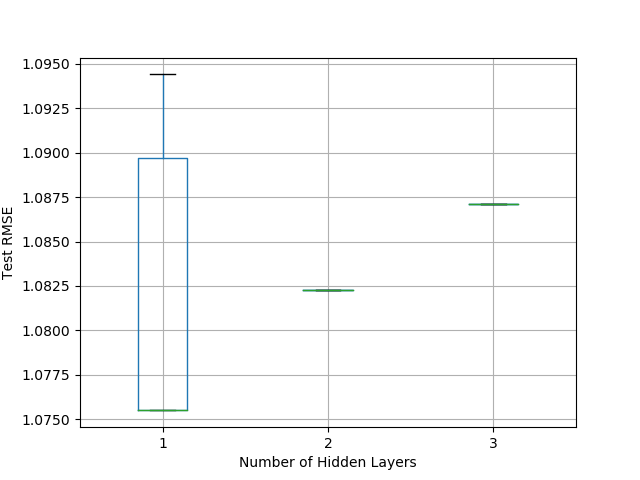}}
	\hfill
	\subfigure[Time Elapsed]{\includegraphics[width=0.32\textwidth]{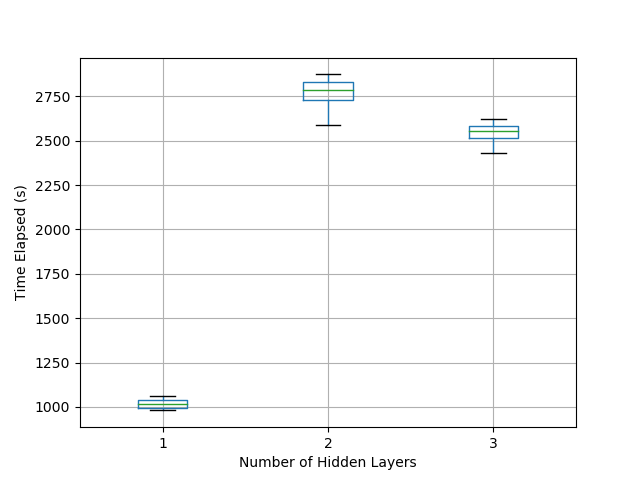}}
	\hfill
	\caption{Method 2 Results, $\alpha=0.10$}
\end{figure}
\FloatBarrier

\subsubsection{Summary of Results}

\begin{table}[!htbp]
	\centering
	\begin{tabular}{|c|c|c|c|}
		\hline
		& 1 & 2 & 3 \\ \hline
		\textbf{Median Score} & 0.92 & 0.93 & 0.93 \\ \hline
		\textbf{Median RMSE} & 1.08 & 1.05 & 1.09 \\ \hline
		\textbf{Median Time Elapsed (s)} & 1014.20 & 9498.86 & 36336.83 \\ \hline
	\end{tabular}
	\caption{Summary of Results for Method 1}
	\label{mnist_benchmark}
\end{table}
\FloatBarrier

\begin{table}[!htbp]
	\centering
	\begin{tabular}{|c|c|c|c|}
		\hline
		& 1 & 2 & 3 \\ \hline
		\textbf{Median Score} & 0.92 & 0.92 & 0.93 \\ \hline
		\textbf{Median RMSE} & 1.08 & 1.08 & 1.09 \\ \hline
		\textbf{Median Time Elapsed (s)} & 1011.10 & 2763.15 & 2556.78 \\ \hline
	\end{tabular}
	\caption{Summary of Results for Method 2, $\alpha= 0.01$}
	\label{mnist_alpha0.01}
\end{table}
\FloatBarrier

\begin{table}[!htbp]
	\centering
	\begin{tabular}{|c|c|c|c|}
		\hline
		& 1 & 2 & 3 \\ \hline
		\textbf{Median Score} & 0.92 & 0.92 & 0.93 \\ \hline
		\textbf{Median RMSE} & 1.08 & 1.08 & 1.09 \\ \hline
		\textbf{Median Time Elapsed (s)} & 1019.95 & 2764.92 & 2539.61 \\ \hline
	\end{tabular}
	\caption{Summary of Results for Method 2, $\alpha= 0.05$}
	\label{mnist_alpha0.05}
\end{table}
\FloatBarrier

\begin{table}[!htbp]
	\centering
	\begin{tabular}{|c|c|c|c|}
		\hline
		& 1 & 2 & 3 \\ \hline
		\textbf{Median Score} & 0.92 & 0.92 & 0.93 \\ \hline
		\textbf{Median RMSE} & 1.08 & 1.08 & 1.09 \\ \hline
		\textbf{Median Time Elapsed (s)} & 1016.91 & 2786.68 & 2553.89 \\ \hline
	\end{tabular}
	\caption{Summary of Results for Method 2, $\alpha= 0.10$}
	\label{mnist_alpha0.10}
\end{table}
\FloatBarrier

\section{Conclusion and Future Work}

The main takeaway from the results obtained is the significant reduction in median time taken to run a test case with an almost equally-high median score and equally-low RMSE when compared to the benchmark case, when the heuristic is used in Method 2. To the best of our knowledge, such a heuristic with the result of a grid search as input has not been properly documented and experimented with, though it is highly possible that it has been formulated and implemented by others given its simple yet seemingly naive nature. 

The heuristic can be generalized and applied to other hyperparameters in a similar fashion, and other models may be used as well. We use the \code{MLPRegressor} and \code{MLPClassifier} models in Scikit-Learn as we find that they help to illustrate the underlying idea of the algorithm the clearest. Due to time constraints we are not able to run for other models and alphas, but we strongly encourage others to explore with other models and variants of the heuristic. 

\bibliographystyle{unsrt}  

\bibliography{references}

\end{document}